\documentclass{article}
\usepackage[utf8]{inputenc}
\usepackage{amsmath}
\usepackage{amsthm}
\usepackage{amssymb}
\usepackage{float}
\usepackage{caption}
\usepackage{subcaption}
\usepackage[]{algorithm2e}
\usepackage{natbib}
\usepackage{xcolor}
\title{Improving neural network robustness through neighborhood preserving layers}
\author{Bingyuan Liu$^{1,2}$, Christopher Malon$^1$, Lingzhou Xue$^2$, Erik Kruus$^1$\thanks{Corresponding author. Email: kruus@nec-labs.com. Address: NEC Laboratories America, Inc. 4 Independence Way, Suite 200 Princeton, NJ  08540}\\ $^1$ NEC Laboratories America, Inc, $^2$Pennsylvania State University}
\date{}
\newcommand{\norm}[1]{\left\lVert#1\right\rVert}
\newcommand{\bx}{{\bf {x}}}
\newcommand{\by}{{\bf {y}}}
\newcommand{\bz}{{\bf {z}}}

\newcommand{\bu}{{\bf {u}}}
\newcommand{\bw}{{\bf{w}}}
\newcommand{\bX}{{\bf {X}}}

\newcommand{\bW}{{\bf W}}

\newtheorem{definition}{\noindent D{\footnotesize EFINITION}}
\newtheorem{assumption}{\noindent A{\footnotesize SSUMPTION}}
\newtheorem{theorem}{\noindent T{\footnotesize HEOREM}}

\newtheorem{lemma}{\noindent L{\footnotesize EMMA}}
\newtheorem{coro}{ \noindent C{\footnotesize OROLLARY}}
\newtheorem{remark}{\noindent R{\footnotesize EMARK}}
\usepackage{natbib}
\usepackage{graphicx}
\usepackage{comment}

\begin{document}
\maketitle

\begin{abstract}
Robustness against adversarial attack in neural networks is an important research topic in the machine learning community. We observe one major source of vulnerability of neural nets is from overparameterized fully-connected layers. In this paper, we propose a new neighborhood preserving layer which can replace these fully connected layers to improve the network robustness. We demonstrate a novel neural network architecture which can incorporate such layers and also can be trained efficiently. We theoretically prove that our models are more robust against distortion because they effectively control the magnitude of gradients. Finally, we empirically show that our designed network architecture is more robust against state-of-art gradient descent based attacks, such as a PGD attack on the benchmark datasets MNIST and CIFAR10.
\end{abstract}

\textbf{keywords:} Deep Learning; Adversarial Attack; Manifold Approximation; Robustness; Neighborhood Preservation; Image Classification.
\section{Introduction}
In the past decade, significant research in machine learning has focused on designing deep neural network architectures for superior prediction performance, especially in visual and natural language processing problems. In a deep neural net, a small perturbation of original data can have a significant effect on the prediction result because of the accumulation effect through overparameterized layers. This fact makes neural network models not robust against adversarial attack. If we carefully design some 'adversarial examples' that are slightly modified from the original sample, the network is likely to misclassify these examples\citep{goodfellow2014explaining,szegedy2013intriguing}. Many works have been established to design algorithms to find imperceptible perturbations to fool the neural networks\citep{moosavi2016deepfool,dong2018boosting,madry2017towards}. It is a major topic in machine learning and computer vision to study how to design networks to be robust against such adversarial attacks.

Adversarial attack methods can be mainly divided into black-box attacks and white-box attacks, based on whether the attack approach has access to the model and model parameters\citep{chakraborty2018adversarial}. White-box attacks are known to be stronger since the attacker gets access to the gradients and model parameters\citep{moosavi2016deepfool,dong2018boosting,madry2017towards,chen2018ead}. Most white-box attacks generate adversarial examples based on the gradient of the loss function with respect to the input. Some representative white-box attacks are the C$\&$W attack\citep{carlini2017towards} and the projected gradient descent (PGD) attack\citep{madry2017towards}. In the C$\&$W attack, the proposed objective function aims at decreasing the probability of the correct class and minimizing the distance between the adversarial example and the original input image. They calculate the gradients with respect to data, and update the data towards the direction of decreasing the probability of correct classification.  The PGD attack directly controls the distortion level by imposing an $\ell_{\infty}$ norm bound $\epsilon$ on the possible change for each input. Because the distortion level can be controlled, the PGD attack is a fair attack method to evaluate the robustness of models against adversarial attack. We will apply this attack approach in this paper to evaluate the performance of different models.

Many theoretical and empirical studies investigate the reason of networks' vulnerability to adversarial attack\citep{hein2017formal,weng2018evaluating,madry2017towards}. These works indicate the large scales of gradients are the major source of vulnerability of networks. By the effect of large scales of gradients, a tiny perturbation in the original data can have a huge effect on the final prediction layer\citep{goodfellow2014explaining}. Especially when the dimensionality of network weights is high, such large scales of gradients are significant. These large gradients usually appear in fully connected layers near the final prediction, serving as dimension reduction layers. For example, in the state-of-art Resnet\citep{szegedy2017inception} or VGG\citep{simonyan2014very} models, we usually encode the image into a $1000$ or so dimensional embedding through convolutional layers, and then need several fully connected layers to reduce the dimension of the embedding from $\sim 1000$ to $\sim 10$ (the number of classes) for final prediction. These fully connected layers have a huge number of weight parameters, and are one major source of the large scale of gradients. For example, in a standard 2-layer CNN network for MNIST dataset, convolutional layers only have less than 20000 parameters, but fully-connected dimension reduction layers have more than 1 million parameters. Such overparameterization is the major source of scales of gradients. If we can control the scale of gradients in such dimension reduction layers, the robustness of neural nets can be significantly improved. 

Based on this observation, we introduce a new neighborhood preserving layer that can replace these fully connected layers, achieving dimension reduction and maintaining all other parts of the neural nets. It has comparable prediction performance but a much smaller scale of gradients. Our proposed neighbor preserving layers share the same input and output dimension as the fully connected layers they replace, thus serving as dimension reduction layers toward the final output. Their property of neighborhood preservation is the key to improving robustness. The neighborhood preservation property guarantees that a small perturbation in any direction will not dramatically change the neighborhood structure, thus will not change the prediction dramatically. More specifically, with same level of attack, our layer has smaller expectation of distortions on the output than fully-connected layers. We verify this property of our layer theoretically and empirically. 

To construct a neighborhood preserving layer, we need to evaluate the quality of neighborhood preservation between inputs and outputs. In literature, many neighbor preserving based dimension reduction methods are related to this topic. Local Linear Embedding(LLE)\citep{roweis2000nonlinear} learns the weights of local neighbors for each point, and then find the optimal low-dimensional embedding in aspect of reconstruction. Neighborhood Preserving Embedding(NPE)\citep{he2005neighborhood} learns linear projections of original data such that the neighborhood structure are most maintained. Isomap\citep{tenenbaum2000global} learns the geodesic distance between data points, and construct embeddings such that reconstruct the distance best. In recent years, UMAP\citep{mcinnes2018umap} and t-SNE\citep{maaten2008visualizing} are the two most popular ones. Both methods have high-dimensional data input, and output a corresponding low-dimensional embedding that can largely maintain the neighborhood structure. These two algorithms have similar ideas. They optimize the low-dimensional embedding such that it minimizes the discrepancy between the neighborhood graphs of the high- and low-dimensional embeddings. The main difference of the two methods is their loss function. The loss function of t-SNE is motivated from a likelihood function, while UMAP is motivated from a uniform manifold assumption, constructing a fuzzy set of the neighborhood graph and then evaluating the discrepancy between the high- and low-dimensional fuzzy sets as the loss function. In this paper, we build our model based on the UMAP loss function, because UMAP is faster and performs better when the dimension of the generated embedding is higher than 2. In Section 3, we will explicitly explain how we incorporate this UMAP loss function into our neighborhood preserving layer.

This paper is organized as follows. Section 2 introduces the details of the model setup, the adversarial attack setup, and background knowledge of neighborhood preserving methods. Section 3 introduces a novel neighborhood preserving layer. The proposed layer can replace fully connected layers in a general classification network, and improve the robustness of a network. We also introduce how to effectively train this network, and combine it with other adversarial training methods. Section 4 theoretically proves why our proposed method is more robust against adversarial attack. Section 5 empirically evaluates the performance of our model on two benchmark datasets: MNIST and CIFAR10. Section 6 concludes our paper.

\section{Model setup and background}
In this section, we introduce the general neural nets and adversarial attack setup, and also relevant dimension reduction methods. In this paper, we will focus on a general setup of neural nets, which is generally applied to most of the state of the art neural network models for prediction. It is worth mentioning that our proposal is also flexible to be applied to other models with internal bottleneck layers under the same philosophy.

\subsection{Model setup}
We consider a general network which can be divided into three parts as follows. \textbf{(1) Encoder:} First the original data is fed into an encoder network, and the output of the encoder is a high-dimensional embedding. The encoder network is usually a multi-layer CNN network for vision tasks with a flattening layer, such as Resnet or VGG. \textbf{(2): Dimension reduction}: Then we pass the data to a dimension reduction layer, where the input dimension is high and output dimension is low; we refer to it as the low dimensional embedding in this paper. It is usually one or multiple fully connected layers. \textbf{(3): Classifier:} Finally, the low-dimensional embedding is a good representation of the data, and can be fed to a small MLP network to determine the final prediction. The final output $\by$ is a vector, whose $c^{th}$ component represents the probability of taking the $c^{th}$ class, and all components sum up to 1.

This network setup fits general classification neural network models, especially for vision tasks. After the encoder, the dimension is usually high. Dimension reduction layers are necessary to map this high-dimensional embedding to a final low-dimensional prediction. For these three steps, we denote the encoder, dimension reduction and classifier functions as $g_1(\cdot)$, $g_2(\cdot)$ and $g_3(\cdot)$ separately; denote input data as ${\bx_i, i=1,\dots,n}$; let corresponding final prediction vectors $\by_i$ represent the probability that data belongs to each class; and let the correct class label of $\bx_i$ be $c(i)$. Then our model can be represented as
$$
\by_i=f(\bx_i) = g_3(g_2(g_1(\bx_i))),
$$
and we aim at minimizing the classification log likelihood loss(or cross entropy)
$
loss = -\sum_{i=1}^n log(y_{ic(i)}),
$
where $y_{ic(i)}$ is the probability that point $\bx_i$ is correctly classified.

In a typical network, $g_2(\cdot)$ is identified as the first one of several fully connected layers\citep{szegedy2017inception,simonyan2014very}. Because a high-dimensional fully connected layer has a large number of parameters, these layers are very vulnerable to adversarial attack, especially to gradient-based adversarial attacks, such as the state of the art PGD attack.

\subsection{Adversarial attack}
Adversarial attacks aim to perturb natural samples without being discovered by humans, but causing a neural net to misclassify the resulting samples. In this paper, we will mainly consider the PGD attack, which is a state of the art white-box adversarial attack model. PGD attacks work as follows. For each data sample, the attack updates the perturbation direction over a certain number of iterations. In each iteration, the PGD attack finds the direction that decreases the probability of the original class most, and then projects the result back to the $\epsilon$-ball of the input in $\ell_{\infty}$ norm. Specifically, for data point $\bx$ in the $t+1^{st}$ iteration, the PGD attack considers moving the original data towards the direction with largest gradient:
$$
\bx^{t+1}=\Pi_{\bx+\epsilon}\{\bx^t+\alpha\text{sign}(\nabla_{\bx}\ell(f(\bx^t),y_0))\}
$$ 
where $\Pi_{\bx+\epsilon}\{\cdot\}$ represents the projection (clipping) into an $\epsilon$ ball in $\ell_{\infty}$ norm, centered at point $\bx$. It makes perturbations with magnitude $\epsilon$. 

\subsection{Dimension reduction with neighborhood preservation}
Here we briefly introduce dimension reduction algorithms which aim at preserving neighborhood structure, such as UMAP and t-SNE, which are closely related to our model.

UMAP and t-SNE are two state-of-the-art unsupervised dimension reduction methods, which can effectively preserve the neighborhood structure of a set of points. The two methods share lots of similarity, and their key idea can be summarized as the following two steps: (1) Construct a neighborhood graph to quantify the membership strength between different points in the input (high dimensional embedding). (2) Find a low dimensional embedding of these points such that its neighborhood graph is most similar to that of the input.

Both methods provide a way to evaluate the similarity of neighborhood graphs. Here we consider UMAP. In step one, we compute the strength between points based on whether they are each others' nearest neighbors and their distances.  Doing this for high- and low-dimensional representations of points yields membership strengths $\mu$ and $\nu$ respectively. Denote by A the set of all unordered pairs of points. For each pair, their membership strength is a value between 0 and 1, representing how close are they in distance. Then in step two, we evaluate the cross entropy C of two fuzzy sets $(A,\mu)$ and $(A,\nu)$:

\begin{small}
$$
C((A,\mu),(A,\nu))=\sum_{a\in A}(\mu(a)\log(\frac{\mu(a)}{\nu(a)})+(1-\mu(a))\log(\frac{1-\mu(a)}{1-\nu(a)}))
$$
\end{small}

The low dimensional embedding which minimizes this cross entropy is found using stochastic gradient descent. 

The two methods provide effective ways to evaluate the similarity between neighborhood structure in high- and low-dimensional embeddings. Typical usage involves a fixed set of input points. In Section~\ref{sec:UmapLayer}, we describe how we extend UMAP to work as a \textit{trainable} neighborhood-preserving dimension reduction layer, $g_2(\cdot)$. In section~\ref{sec:Theory} we consider the properties of such a mapping from a theoretical perspective, followed by empirical demonstrations in section~\ref{sec:Experiments}.

\section{A novel neighborhood preserving layer}
\label{sec:UmapLayer}
\subsection{Network structure}
In this section, we propose a novel neighborhood preserving layer that can replace fully connected layers in dimension reduction, which can significantly improve network robustness. 

As we discuss, the dimension reduction part($g_2(\cdot)$) is a major source of vulnerability. If we can bridge the high-dim embedding and low-dim embedding without introducing a dimension reduction layer with large scale gradients, the robustness can be significantly improved. The proposed layer serves as a new $g_2(\cdot)$, and replaces one or many fully-connected layers for dimension reduction in a general classification neural network introduced in Section 2. 

To achieve neighborhood preservation in dimension reduction layer, the key is to find a nice representation of low-dimensional embedding $g_2(g_1(\bx))$ satisfying the following conditions: (1) \textbf{Neighbor Preservation}: It preserves the neighborhood structure of the high-dimensional embedding $g_1(\bx)$. (2) \textbf{Precise Prediction}: It achieves a good prediction performance after the classifier $g_3(\cdot)$, i.e. $g_3(g_2(g_1(\bx)))$ is a good prediction. Therefore it is important to establish criteria to evaluate both neighborhood preservation and prediction error.

For condition (2) on prediction, it can be evaluated by the standard log likelihood loss function as introduced in Section 2.1. For condition (1) on neighborhood preservation, we need to introduce a metric to evaluate the discrepancy between the neighborhood structure of the high dimensional embedding and low dimensional one. This type of metric already is used in UMAP and t-SNE algorithms. Here we follow the spirit of dimension reduction methods, and propose a metric to evaluate neighborhood preservation through the similarity between two neighborhood graphs. We introduce explicit steps to compute this similarity metric, following the loss function of the UMAP algorithm in \citep{mcinnes2018umap}. We denote $A$ as the set of all pairs between $1,\dots,n$, who are neighbors of each other. For any unordered pair $(i,j) \in A$, $\mu(i,j)$ represents the membership strength of the pair $(\bx_i,\bx_j)$. Membership strengths $\mu(i,j)$ are computed in the following steps:
\begin{itemize}
    \item For each point $x_i$, we search its $k$ nearest neighbors, and denote the distance to these neighbors as $d_{i1},\dots,d_{ik}$. Then we normalize these distances as: $d^{'}_{ij}=d_{ij}-\underset{t=1,\dots,k}{\min}\{d_{it}\}$.
    \item Compute the ordered membership strength between $x_i$ and its neighbor $x_j$ as:
    $z_{ij}=e^{-d^{'}_{ij}}$.
    \item For each pair $(i,j) \in A$, the unordered membership strength is computed as the average of two ordered ones $z_{ij}$ and $z_{ji}$: $\mu(i,j)=\frac{z_{ij}+z_{ji}}{2}$.
\end{itemize}
For the ease of notation, the previous procedure of computing membership strengths is denoted by the function $\mathcal{M}(\cdot)$, i.e. $(A,\mu)=\mathcal{M}(g_1(\bx))$. \citep{mcinnes2018umap} provides complete theoretical justification for this procedure of computing membership strength.

 Equipped with these metrics, we propose a novel neighborhood preserving layer in a general classification network. Instead of using a fully connected layer as $g_2(\cdot)$, we don't assume a one-to-one continuous parameterized function as $g_2(\cdot)$ here. Instead, we store a table that maps a reference set of high dimensional points to their low dimensional embeddings. We consider this reference set to be obtained from training data, so their high dimensional point is $\bu^i_{high}=g_1(\bx_i)$ for a set of data point indices $\{i\}$. Their corresponding low-dimensional embeddings $\bu_{low}$ are stored and learned.  We store a mapping $G_2:i\mapsto \bu^i_{low}$, over the fixed set of inputs $\{i\}$. In this aspect, $\bu^i_{low}$ can be treated as the parameters of our layer, and we update them through back propagation, with respect to the full loss function. In the back propagation, we back propagate the derivative with respect to parameters in encoder(CNN) and classifier(purple arrows in Figure 1), and also the low dimensional embeddings $\bu_{low}$.

\begin{figure}[H]
    \centering
    \includegraphics[width=3.5in]{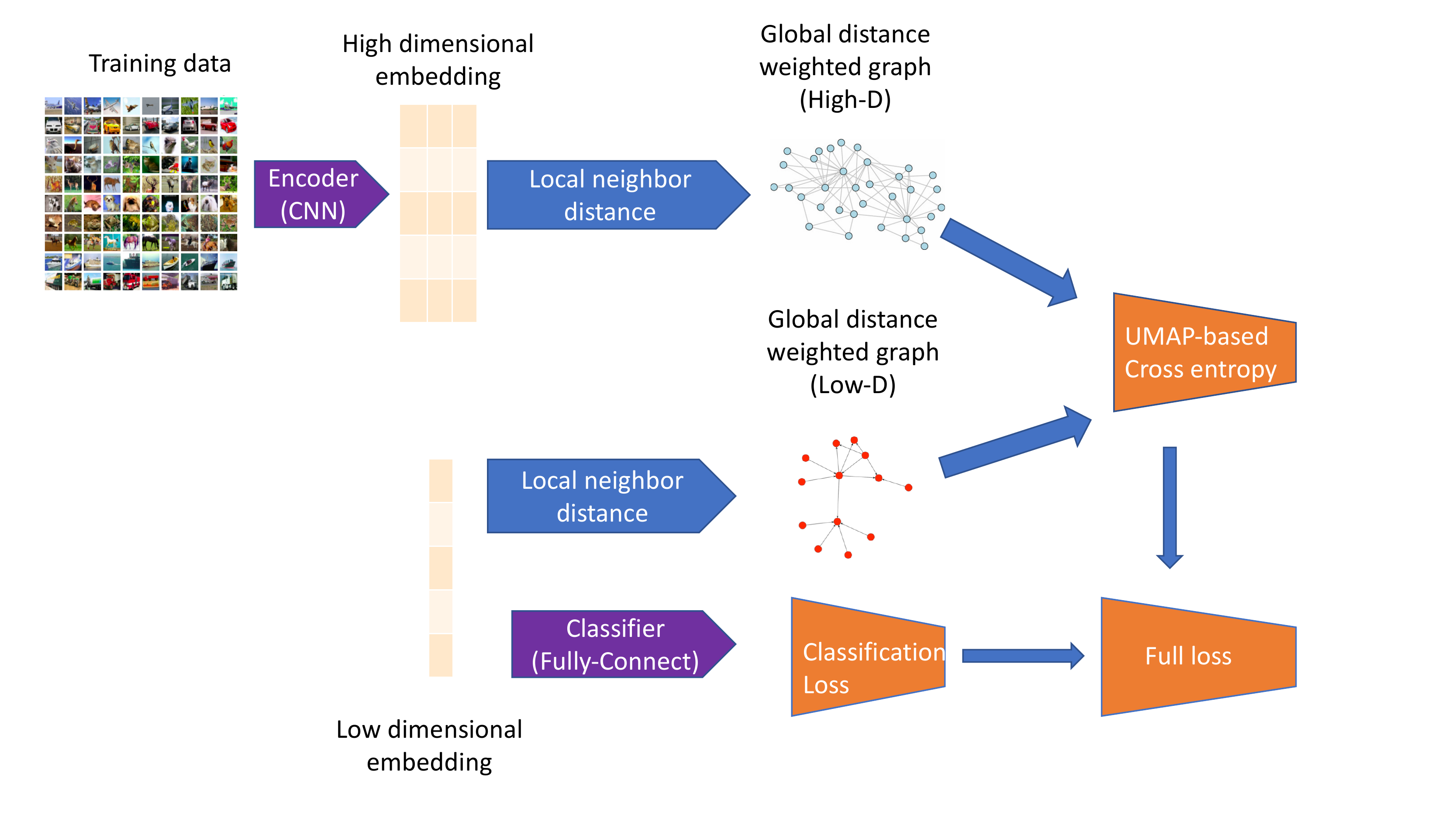}
    \caption{Training network architecture}
    \label{fig:train_plot}
\end{figure}

We visualize the pipeline of our proposed training network in Figure \ref{fig:train_plot}. The explicit procedure of all steps are summarized as follows. 
\begin{itemize}
\item \textbf{Encoder(CNN)}: $\bu_{high}=g_1(\bx)$
\item {\textbf{Local neighbor distance(High-D)}:$(A,\mu)=\mathcal{M}(\bu_{high}$)}
\item {\textbf{Local neighbor distance(Low-D)}:$(A,\nu)=\mathcal{M}(\bu_{low}$)}
\item {\textbf{Cross entropy loss: $L_G =\sum_{a\in A}(\mu(a)\log(\frac{\mu(a)}{\nu(a)})+(1-\mu(a)\log(\frac{1-\mu(a)}{1-\nu(a)}))
$}}
\item \textbf{Classification loss}: $L_C = -\sum_{i=1}^n log(y_{ic(i)})$
\item {\textbf{Full loss: $L_{full} = L_C + \alpha L_G$}}
\end{itemize}

In Figure \ref{fig:train_plot}, the blue and purple arrows show the data flow for forward propagation. The key difference between a general classification network and our proposed network is that, during the back-propagation, our low-dimensional embedding is \textit{not} viewed as a parameterized function $g_2(\cdot)$ of the high-dimensional embedding. There is no data flow arrow directly from the high-dimensional to low-dimensional embedding $\bu_{high}\mapsto \bu_{low}$. Instead, each embedding is independently updated through back-propagation to maintain the neighborhood structure of the high-dimensional embedding as much as possible, and also achieve good classifier predictions.

By combining the classification loss and neighborhood preserving loss, we update the parameters to achieve a good classification result and maintain the consistency between neighborhood graphs of the high-dimensional and low-dimensional embeddings at the same time. The proposed algorithm also needs an initialization of the low-dimensional embedding for each data point to start training. A reliable initialization can be obtained by performing the UMAP algorithm on the high-dimensional embedding. We stop the back propagation when the loss function converges in the training procedure.

After training, we obtained trained weights of the encoder and classifier, and also the low-dimensional embedding for all points in the training data. Here we call the low dimensional embedding for all training data as 'reference points', because they can be the reference for us to predict unseen data. Then we discuss how we can predict unseen new points. They cannot be fed into the training network directly, because these unseen points don't have a low dimensional embedding yet. Therefore, we need a method that can make use of the information of the mapping $G_2:i\mapsto \bu^i_{low}$ in the training set, to obtain the low dimensional embedding for unseen points. Here we use a weighted k-nearest neighbors approach. For each unseen point $\bx$, the complete forward propagation procedure is summarized as follows:

\begin{itemize}
    \item Compute high dimensional embedding for new point $\bx$: $\bu_{high}=g_1(\bx)$
    \item Find k-nearest neighbors of $\bu_{high}$ from reference points: $\bu^1_{high},\dots,\bu^k_{high}$. Suppose the corresponding low-dimensional embedding of these $k$ reference points are $\bu^1_{low},\dots,\bu^k_{low}$.
    \item Compute the low dimensional embedding as $\bu_{low}=(\sum_{i=1}^kw_i\bu^i_{low})/(\sum_{i=1}^kw_i)$. Weight $w_i$ is a decreasing function of distance from $\bu_{high}$ to $\bu^i_{high}$.
    \item Feed this low-dimensional embedding to the classifier to obtain the final prediction.
\end{itemize}

The complete prediction pipeline is displayed in Figure 2. Our weights computation in prediction is consistent with training framework, thus provide us an accurate prediction on whether the point would be if it is in the training data. Comparing with the out-of-sample prediction for LLE and ISOmap\citep{bengio2004out}, we emphasize the pair-wise distance thus further guarantee the robustness against perturbations.

\begin{figure}[H]
    \centering
    \includegraphics[width=3.5in]{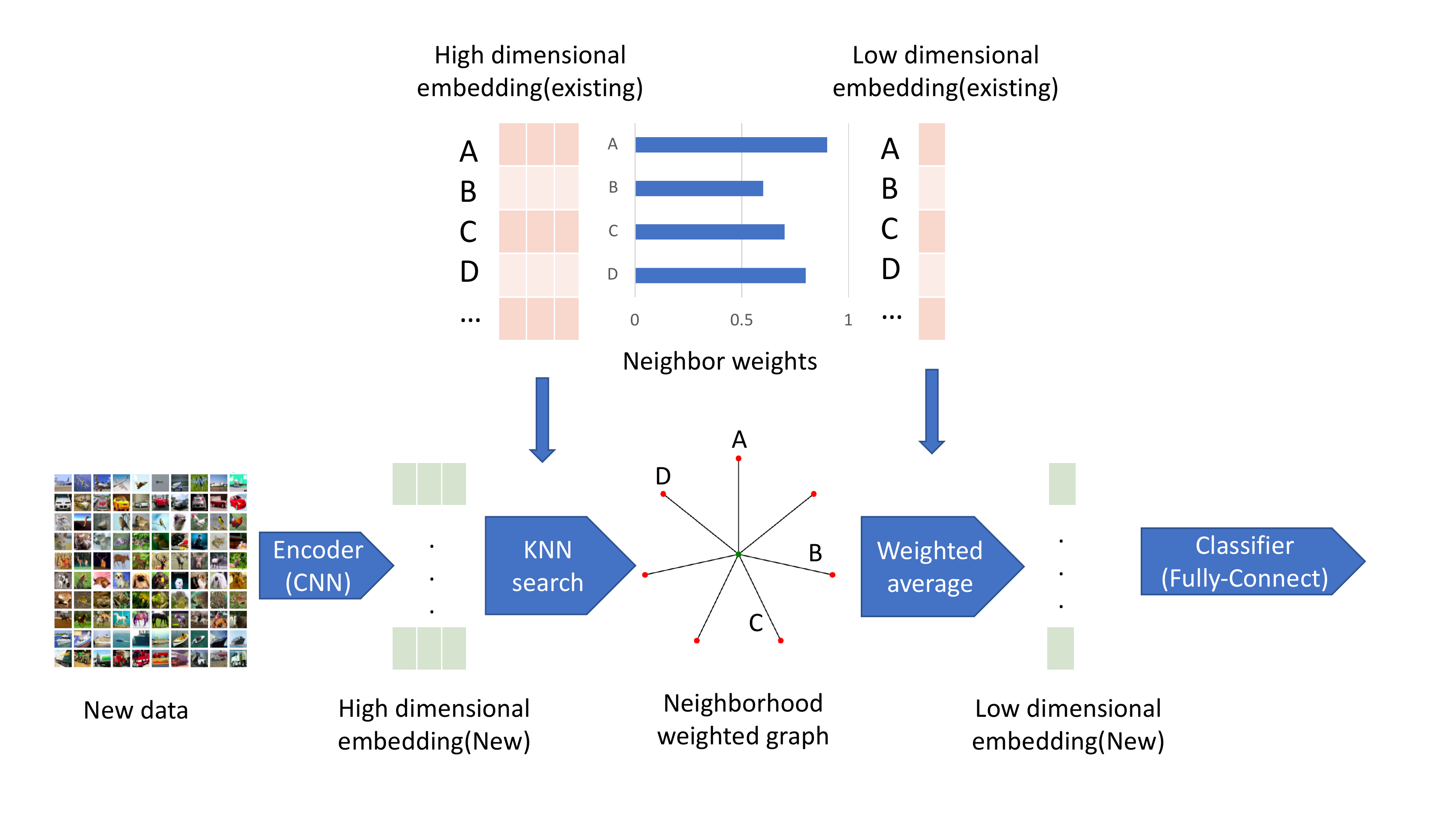}
    \caption{Prediction network architecture. Here A,B,C,D are neighboring \textit{reference points} that establish an interpolated mapping of $\bu_{high}\mapsto \bu_{low}$.}
\end{figure}

Instead of imposing regularizations\citep{ross2017improving} or dropout\citep{srivastava2014dropout} on fully-connected layers, Our proposed layer fundamentally reduce the number of parameters in our network, such that it boosts the robustness of our models. 

\subsection{Adversarial training}
Our framework can also adapt standard adversarial training approaches, to further improve the robustness against adversarial attack. In adversarial training, for each batch of training data, we generate an `adversarial batch' through the attack method, such as PGD attack. Then we assume the labels of adversarial batch are the same as in the training batch, and we train the model weights on both the original batch and the adversarial batch. In this way, the network will be less sensitive to the attack.

\begin{figure}[H]
    \centering
    \includegraphics[width=3.5in]{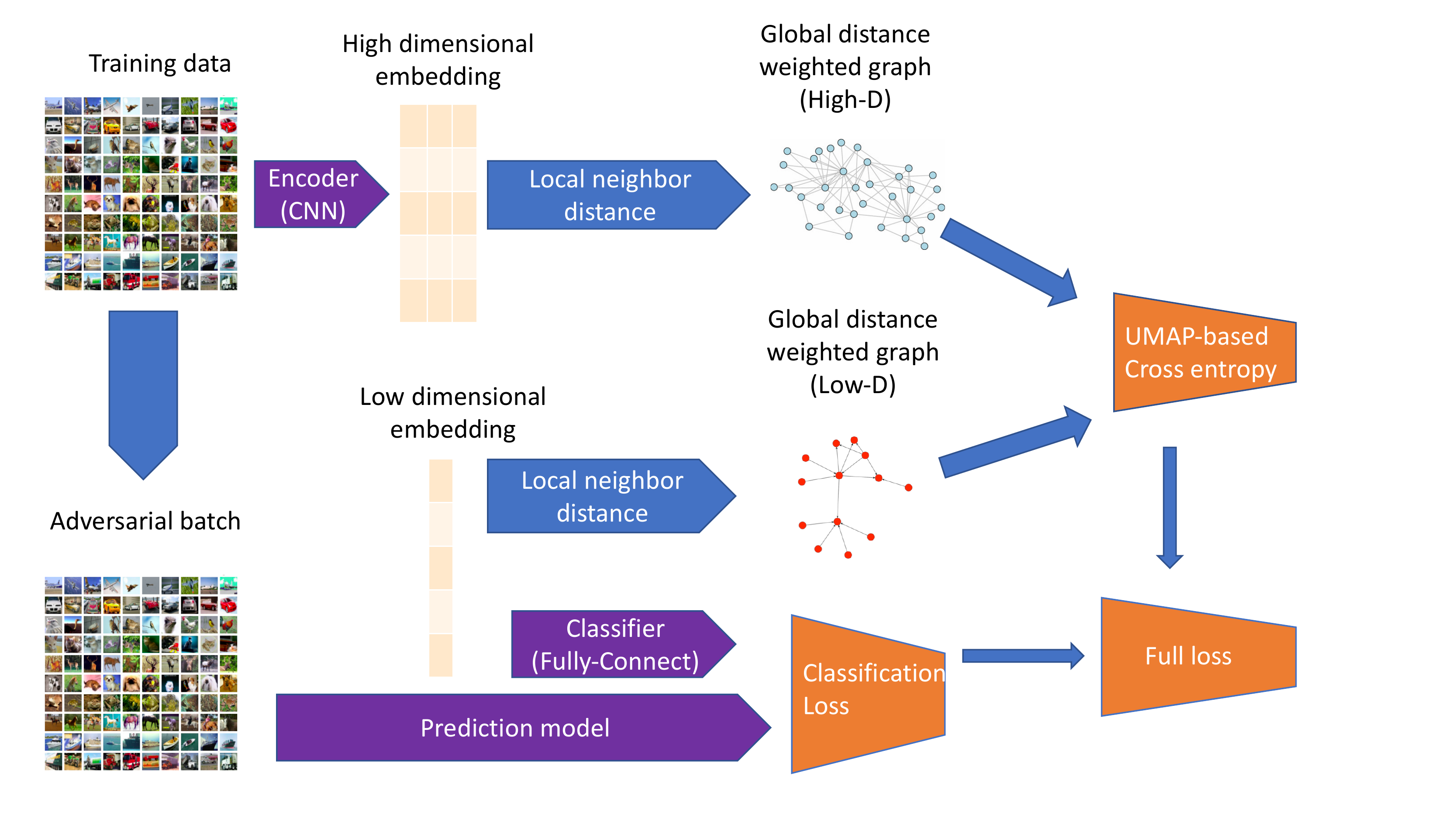}
    \caption{Adversarial training network architecture.}
    \label{fig:train_adv_plot}
\end{figure}

This procedure also works for our network. With each training batch, the adversarial batch can also be generated using the PGD attack algorithm. The prediction of an adversarial batch is forward propagated using the prediction procedure as introduced in Section 2.1. Because the prediction procedure is completely differentiable, the adversarial batch can be generated and forward propagated smoothly. Finally, we evaluate the loss function on both the true data and a generated `adversarial batch'. The pipeline of the adversarial training procedure is shown in Figure \ref{fig:train_adv_plot}. 

\subsection{Using representative points }
\label{subsec:RefPoints}

In a more general case, $G_2:\bu^r_{high}\mapsto \bu^r_{low}$ for $r=1\ldots R$.  This might happen when input data includes a potentially infinite stream of input points (like data augmentation).  We store this and learn a more general table, but it might cost a lot of memory to store the domain $\{\bu^r_{high}\}$. We might save storage by using \textit{representative points} that have occurred as actual inputs $\{\bx_r\}$ and recalculating $\bu^r_{high}=g_1(\bx_r)$ as needed. In this case, we need to construct the set of representative points.

We introduce a heuristic approach to construct a smaller set of representative points. First, we train the network as same as proposed in Section 3.1. Using all high-dimensional embeddings, we use the K-means clustering algorithm with 100 clusters inside each class. For the 10-class MNIST dataset, we then have 1000 clusters with cluster centers $\bx_i$ and $\bz_i$ in high and low dimensions. The center of high/low dimensional embedding of these 1000 clusters are the `representative points'.  Each cluster has weight $v_i$, which is the size of the cluster. This reference point set with size 1000 is a good representation of all points. By such properly selected representative points, we can largely reduce the computational burden and maintain a good prediction performance.

The size of representative points can be further reduced if proper points can be selected. For example, \citep{afsar2008pruned,arif2010pruned} propose a way to choose representative points based on the exterior of well-separated clusters.

\section{Theoretical analysis}
\label{sec:Theory}
In this section, we illustrate the theoretical advantage of our proposed network structure. We prove that our proposed dimension reduction layer has a smaller gradient upper bound compared with the general neural network structure with fully-connected layers. We show that the decreasing of the gradient bound leads to significant improvement of network robustness, as measured by acknowledged robustness metrics such as minimal $L_p$ distortion.

First let's review current literature. The studies of robustness of neural nets heavily relies on the gradient bounds of networks. In this section, we study the gradient bounds of our proposed model versus standard neural net models, and then study how the bound affect the robustness. Because both models share the same structure of encoder and classifier, we focus on studying the gradient bounds of dimension reduction layers for both models. 

Here we explicitly introduce the setup of our theoretical analysis of the dimension reduction layer. To fairly compare our proposed neighbor preserving layer with a fully connected layer, we assume they have same scale, i.e. both layers map $\bx \in \mathbb{R}^p\mapsto\by\in \mathbb{R}^d$, with boundary points  all mapped to boundary points. Without loss of generality, we assume inputs and outputs are bounded by $\ell_2$ bounds around origin with fixed radius. Fixing the scale will not affect the prediction ability of the network, since the network between layers can be adjusted by a multiplier correspondingly. 

Our proposed layer is based on a mapping of representatives $G_2:\bx_r\in \mathbb{R}^p \mapsto \by_r\in\mathbb{R}^d$ for $r=1\ldots N$. Mapping a generic input $\bx_0$ follows the same neighborhood weighted average approach we used in the prediction network:
$$
f(\bx_0)=\frac{\sum_{i=1}^k w_i\by_i}{\sum_{i=1}^k w_i} \text{ for neighbor representatives }\{\bx_1,\dots,\bx_k\}\text{ of }\bx_0\in\mathbb{R}^p.
$$
It is worth mentioning that, in our analysis, instead of fixing the number of nearest neighbors $k$, we fix the maximum distance between neighbors and point $\bx_0$. We choose all points which are inside $\mathbb{B}_r(\bx_0)$ as the neighbors. We denote the number of points inside $\mathbb{B}_r(\bx_0)$ as $n$ in this section. Choosing the radius of ball $r$ is equivalent to choosing a specific number of neighbors $k$.

We introduce another assumption to bound the neighborhood update frequency. It represents the ratio of points changed as an $\mathbb{B}_r$-ball moves by a small $\epsilon$.
\begin{assumption}
$\bx$ follows a distribution $\mathbb{P}$ with $\|\bx\|_{max} \leq C_0$ for some constant $C_0$. We assume $\mathbb{P}$ is uniformly bounded in density almost everywhere.
\end{assumption}

Furthermore, we introduce a class of distribution beforehand. We say a distribution is an $\alpha$-even distribution if for any two regions with same volume A,B in feasible region $S$, such that for a sample point $x \in P$:
$$
\underset{vol(A)=vol(B)}{\max} \frac{p(x\in A)}{p(x\in B)} \leq \alpha
$$
All uniformly bounded distributions with density almost everywhere can be represented as an $\alpha$-even distribution since their density is both upper and lower bounded.

\begin{lemma}
If $\bx$ follows an $\alpha$-even distribution, then for any fixed point $\bx_0$, then as the sample size $N \to \infty$, with probability going to 1:
{\small
$$
\underset{N \to \infty}{\lim}\underset{\|\epsilon\|_2\to 0}{\lim}\frac{Card_{x}(x\in \mathbb{B}_r(\bx_0), x \notin  \mathbb{B}_r(\bx_0+\epsilon))+Card_{x}(x\in \mathbb{B}_r(\bx_0+\epsilon), x \notin  \mathbb{B}_r(\bx_0))}{N\epsilon} = 0
$$
}
\end{lemma}
\begin{proof} For any $\epsilon$, we can calculate the volume of the intersection of two high-dimensional balls and their symmetric difference. Referring to \citep{li2011concise}, we have:
$$
\frac{vol(\mathbb{B}_r(\bx_0) \ \Delta \  \mathbb{B}_r(\bx_0+\epsilon))}{vol(\mathbb{B}_r(\bx_0))}=\mathbb{I}_{(\epsilon/2r)^2}(\frac{p+1}{2},\frac{1}{2})
$$
where $\Delta$ is the symmetric difference operator $A \Delta B= (A \cap B^c) \cup (A^c \cap B)$, and $\mathbb{I}$ is the regularized incomplete beta function:
$$
\mathbb{I}_{t}(a,b)=\frac{\int_0^rx^a(1-x)^bdx}{\int_0^1x^a(1-x)^bdx}
$$
As $\epsilon \to 0$, we can find a sufficiently small $\epsilon$ such that $x^{\frac{p+1}{2}}(1-x)^{1/2}$ decreases with $x$ on $(1-(\epsilon/2r^2),1)$. Then we know:
$$
\int_{0}^{(\epsilon/2r)^2}x^{\frac{p+1}{2}}(1-x)^{\frac{1}{2}}dx \leq (\epsilon/2r)^2(1-\epsilon/2r)^{\frac{p+1}{2}}((\epsilon/2r)^2)^{\frac{1}{2}} \leq (\epsilon/2r)^{3}
$$

On the other hand, we have that the fraction in $\mathbb{I}_{t}(a,b)$ is a Beta function:
$$
\int_0^1x^a(1-x)^bdx = Beta(\frac{p+3}{2},\frac{3}{2})
$$
 Thus 
\begin{align}
\frac{\int_{1-(\epsilon/2r)^2}^1x^{\frac{p+1}{2}}(1-x)^{\frac{1}{2}}dx}{\epsilon\int_0^1x^{\frac{p+1}{2}}(1-x)^{\frac{1}{2}}dx} \to 0
\end{align}
and  $\mathbb{I}_{1-(\epsilon/2r)^2}(\frac{p+1}{2},\frac{1}{2}) \to 1$ as $\epsilon \to 0$. It means that for any constant $C_3 > 0$ and for all $\bx_0$, we have
$$
\frac{vol(\mathbb{B}_r(\bx_0) \ \Delta \ \mathbb{B}_r(\bx_0+\epsilon))}{\epsilon* vol(\mathbb{B}_r(\bx_0))} < \frac{C_3}{\alpha}
$$
for sufficiently small $\epsilon$. 
For an $\alpha$-even distribution, by definition and the previous inequality, we know for any point $\bx$,
$$
\frac{p(\bx \in \mathbb{B}_r(\bx_0) \ \Delta \ \mathbb{B}_r(\bx_0+\epsilon))}{\epsilon*p(\bx \in vol(\mathbb{B}_r(\bx_0)))} < \frac{C_3}{\alpha}\alpha = C_3
$$

By the law of large numbers, for sufficiently large $N$ (and hence $n$) with probability going to 1 we have:
{\footnotesize
$$
\underset{\|\epsilon\|_2\to 0}{\lim}\frac{Card_{x}(x\in \mathbb{B}_r(\bx_0), x \notin  \mathbb{B}_r(\bx_0+\epsilon))+Card_{x}(x\in \mathbb{B}_r(\bx_0+\epsilon), x \notin  \mathbb{B}_r(\bx_0))}{n\epsilon} < C_3
$$
}
The lemma follows because $C_3$ can be taken to be arbitrarily small.
\end{proof}

Another assumption is that all the points in $\mathbb{B}_r(\bx_0)$ and $\mathbb{B}_r(\bx_0+\epsilon)$ are uniformly bounded in $\ell_2$ norm. 
\begin{assumption}
For point $\bx_0$, we assume for any point $\bx_i \in \mathbb{B}_r(\bx_0)$, its embedding $\by_i$ satisfies $\|\by_i\|_2 \leq C_4$.
\end{assumption}
Since our goal is to obtain a well-behaved low-dimensional embedding, such a bound is reasonable in our setting. 

We also introduce necessary notation. We introduce a scalar distance-weighting function $h$ with derivative $h'$, which can be applied element-wise to vectors as $h(\bz)$. We use $cov_{\bz \in S}(\bz,h(\bz))$ to represent the element-wise population covariance between each element in random vector $\bz$ and a random scalar $h(\bz)$, restricting $\bz$ inside set $S$. This covariance has the same dimension as $\bz$. Further, for a data point $\bx_0$, we assume its neighbors in $\mathbb{B}_r(\bx_0)$ are $\bx_1,\dots,\bx_n$ and their embeddings are $\by_1,\dots,\by_n$. Their distances to $\bx_0$ are denoted as $d(\bx_0,\bx_i)$ for $i=1,\dots,n$. In practice, we use $w_i = h(d(\bx_0,\bx_i)) = \exp(-d(\bx_0,\bx_i))$ as our weight function.

\begin{theorem}
Suppose data follows distribution $\mathbb{P}$ in space $\mathbb{R}^p$, and we uniformly sample $N$ points from $\mathbb{P}$.  Take all points inside $\mathbb{B}_r(\bx_0)$ as the reference points of $\bx_0$. Take $C_1$ and $C_2$ such that $E_{\bz\in \mathbb{B}_r(\bx_0)}h(d(\bx_0,\bz)) \geq C_1$ and $E_{\bz\in \mathbb{B}_r(\bx_0)}|h^{'}(d(\bx_0,\bz))| \leq C_2$. Assume assumption 1 and 2 apply with constant $C_4$. Given $\delta > 0$, for all direction vector $\bu$ normalized so that $\norm{\bu}=1$, we have for $N$ sufficiently large:
$$
\underset{c\to 0}{\lim} \|\frac{f(\bx_0+c\bu)-f(\bx_0)}{c}\|_2 \leq  2C_4(\frac{C_2}{C_1})+\delta
$$
with probability going to 1.
\end{theorem}

\begin{remark}
From the theorem, we see that the Lipschitz bound of neighborhood weighted embedding is determined by $(C_1,C_2,C_4,\alpha)$. By the definition, we know $C_1 \leq 1$ by our choice of radius $r$; $C_2$ is lower bounded as long as we choose a function decay sufficiently fast. For example, if we choose $h(x)=exp(-x)$, we have $C_2 \leq 1$; $C_4$ is also a small constant independent of $p$.

Therefore the Lipschitz bound of our neighborhood embedding layer is smaller and will not diverge with $p$, and is free from the scale of $\bx$.
\end{remark}

\begin{remark}
Connection with intrinsic dimension: The theorem results also implicitly connect with the intrinsic dimension of $\bx$. Here we take a sufficient large $N$, but actually this $N$ is connected with the intrinsic dimensionality of $\bx$. From (Levina and Bickel 2005), we know if the intrinsic dimension of $\bx$ is $m$, then the Euclidean distance from a fixed point $\bx$ to its $k$-th nearest neighbor $T_k(\bx)$ approximately satisfies:
$$
\frac{k}{N} \approx f(\bx)(T_k(\bx))^mV(m)
$$
where $V(m)=\frac{\pi^{m/2}}{\Gamma(m/2+1)}$, is the volume of the unit sphere in $\mathbb{R}^m$.

Here for our neighbor search procedure in Theorem 1, we have $T_k(\bx)=r$ fixed. Therefore for any fixed $N$, the number of neighbors $k$ can be approximated as:
$$
k \approx Nf(\bx)(r)^mV(m) \approx \frac{N}{\sqrt{\pi m}}f(\bx)(2e\frac{\pi r^2}{m})^{m/2}
$$
This is an increasing function of $m$ roughly when $m<2e\pi r^2$. Therefore we know the sample size we need is small when $m$ is not large. It explains why we can achieve a much smaller gradient upper bound with relatively small sample sizes, when the intrinsic dimension is not large. On the other hand, when the intrinsic dimension is large, we should not improve robustness through neighborhood preserving layers, because such low dimensional embeddings are not stable. It also suggests that we should choose $r$ according to the true intrinsic dimensionality. The larger the intrinsic dimension, the larger the $r$ should be.
\end{remark}

\begin{proof}
The proof is separated into two parts. First we consider the derivative w.r.t to $\bx$ if the set of neighbors does not change. Then we consider the case of neighbor change.

First, if there is no neighbor change, we can just consider derivatives w.r.t. every possible high dimensional embedding of $\bx_0$. Denoting $g_i(\bx_0)=\frac{w_i}{\sum_{i=1}^n w_i}$, we can calculate the derivatives in a specific direction $\epsilon$ such that $\|\epsilon\|_2=1$:
\begin{align*}
    &\underset{c \to 0}{\lim}\frac{f(\bx_0+c\epsilon)-f(\bx_0)}{c}\\
    =&\sum_{i=1}^n\by_i\underset{v \to 0}{\lim}\frac{ g_i(\bx+c\epsilon)-g_i(\bx)}{c} \\
    =&\sum_{i=1}^n\by_i\frac{w_i^{'}(\bx)\sum_{i=1}^n w_i - w_i \sum_{i=1}^n w^{'}_i(\bx)}{(\sum_{i=1}^n w_i)^2} \\
    =&\frac{\sum_{i=1}^n\by_iw_i^{'}(\bx)}{\sum_{i=1}^nw_i}-\frac{\sum_{i=1}^nw_i^{'}(\bx)\sum_{i=1}^n\by_iw_i}{(\sum_{i=1}^n w_i)^2} \\
        =&\frac{\sum_{i=1}^n\by_iw_i^{'}(\bx)/n}{\sum_{i=1}^nw_i/n}-\frac{\sum_{i=1}^nw_i^{'}(\bx)/n\sum_{i=1}^n\by_iw_i/n}{(\sum_{i=1}^n w_i/n)^2} 
        \end{align*}
where $w_i^{'}(\bx)=\underset{c \to 0}{\lim}\frac{w_i(\bx+c\epsilon)-w_i(\bx)}{c}$, is the gradient of $w_i(\bx)$ in the direction $\epsilon$. Therefore we can bound its $\ell_2$ norm as:     
        \begin{align*}
           &\|\underset{c \to 0}{\lim}\frac{f(\bx_0+c\epsilon)-f(\bx_0)}{c}\|_2 \\
           \leq & C_4 (|\frac{\sum_{i=1}^n|w_i^{'}(x)|/n}{\sum_{i=1}^nw_i/n}|+|\frac{\sum_{i=1}^n|w_i^{'}(x)|/n\sum_{i=1}^nw_i/n}{(\sum_{i=1}^n w_i/n)^2}|) \\
          \leq & C_4(\frac{\sum_{i=1}^n|h^{'}(d(x_0,x_i))|/n}{\sum_{i=1}^n|w_i|/n}+\frac{\sum_{i=1}^n|h^{'}(d(x_0,x_i))|/n\sum_{i=1}^n|w_i|/n}{(\sum_{i=1}^n |w_i|/n)^2}) \\
           =& 2C_4\frac{\sum_{i=1}^n|h^{'}(d(x_0,x_i))|/n}{\sum_{i=1}^n|w_i|/n}
\end{align*}
where we use the fact that $w_i^{'}(x) \leq h^{'}(d(x_0,x_i))$, and equality holds if and only if $\epsilon$ is exactly in the direction of $(\by_i-\bx)$.


Then we consider the convergence of the empirical average to this expectation. We know  $\lim_{n \rightarrow \infty} \sum_{i=1}^n|h^{'}(d(x_0,x_i))|/n \leq C_2$ and $\lim_{n \rightarrow \infty} \sum_{i=1}^nw_i/n \geq C_1$. Further $w_i$ and $1/w_i$ are all bounded with finite second moment values. Thus we can apply Slutsky's theorem, to obtain:
$$
  \underset{c \to 0}{\lim}\|\frac{f(\bx_0+c\epsilon)-f(\bx_0)}{c}\|_2 \leq  2C_4\frac{\sum_{i=1}^n|h^{'}(d(x_0,x_i))|/n}{\sum_{i=1}^nw_i/n} \overset{P}{\to} 2\frac{C_4C_2}{C_1}
$$
This result indicates that for any $\delta>0$, there exists $n_0$ such that for $n \geq n_0$:
\begin{align*}
 \underset{c \to 0}{\lim}\|\frac{f(\bx_0+c\epsilon)-f(\bx_0)}{c}\|_2\leq \frac{2C_4C_2}{C_1} + \delta
\end{align*}
Indeed, let $p_1$ be the probability that each point falls in this ball. We know with sample size $N$, the number of points falling in the ball follows a binominal distribution $Bin(N,p_1)$. Therefore with probability going to 1, when we have $N\geq 2n_0/p_1$, the number of points falling in the ball satisfies $n>n_0$, and it guarantees that  $\underset{c \to 0}{\lim}\|\frac{f(\bx_0+c\epsilon)-f(\bx_0)}{c}\|_2\leq \frac{2C_4C_2}{C_1} + \delta$ holds.

Now consider the case where neighbors change. We denote by $\bx_i$ for $i=1,\dots,n-k$ the points in both $\mathbb{B}_r(\bx)$ and $\mathbb{B}_r(\bx+\epsilon)$. We denote by $\bx_i$ for $i=n-k+1,\dots,n$ the points in $\mathbb{B}_r(\bx)$ but not $\mathbb{B}_r(\bx+\epsilon)$. We denote by $\bx_i^{new}$ for $i=1,\dots,k'$ the points in $\mathbb{B}_r(\bx+\epsilon)$ but not in $\mathbb{B}_r(x)$. We use $w_i$ to denote the weight with respect to $\bx$, and $w_i^{'}$ as the weight with respect to $\bx+\epsilon$. Then integrating the effect of updating neighbors and updating weights, the embedding change can be bounded as:
{\footnotesize
\begin{align*}
\Delta f(\bx)&=(\frac{\sum_{i=1}^nw_i\by_i}{\sum_{i=1}^nw_i}-\frac{\sum_{i=1}^nw^{'}_i\by_i}{\sum_{i=1}^nw_i^{'}})+(\frac{\sum_{i=1}^nw^{'}_i\by_i}{\sum_{i=1}^nw_i^{'}}-\frac{\sum_{i=1}^{n-k}w^{'}_i\by_i+\sum_{n=1}^{k'}w_i^{new}\by_i^{new}}{\sum_{i=1}^{n-k}w_i^{'}+\sum_{n=1}^{k'}w_i^{new}}) \\
&= (A) + (B)
\end{align*}
}
The (A) part is bounded by the previous gradient bound. We will focus on bounding the (B) part.
\begin{align*}
    (B)&=(\frac{\sum_{i=1}^nw^{'}_i\by_i}{\sum_{i=1}^nw_i^{'}}-\frac{\sum_{i=1}^nw^{'}_i\by_i}{\sum_{i=1}^{n-k}w_i^{'}+\sum_{n=1}^{k'}w_i^{new}})\\
    &+(\frac{\sum_{i=1}^nw^{'}_i\by_i}{\sum_{i=1}^{n-k}w_i^{'}+\sum_{n=1}^{k'}w_i^{new}}-\frac{\sum_{i=1}^{n-k}w^{'}_i\by_i+\sum_{n=1}^{k'}w_i^{new}\by_i^{new}}{\sum_{i=1}^{n-k}w_i^{'}+\sum_{n=1}^{k'}w_i^{new}}) \\
    &=(C)+(D)
\end{align*}

where 
$$(C) = \frac{\sum_{i=1}^nw^{'}_i\by_i}{\sum_{i=1}^nw_i^{'}}-\frac{\sum_{i=1}^nw^{'}_i\by_i}{\sum_{i=1}^{n-k}w_i^{'}+\sum_{n=1}^{k'}w_i^{new}}$$ and $$(D) = \frac{\sum_{i=1}^nw^{'}_i\by_i}{\sum_{i=1}^{n-k}w_i^{'}+\sum_{n=1}^{k'}w_i^{new}}-\frac{\sum_{i=1}^{n-k}w^{'}_i\by_i+\sum_{n=1}^{k'}w_i^{new}\by_i^{new}}{\sum_{i=1}^{n-k}w_i^{'}+\sum_{n=1}^{k'}w_i^{new}}.$$

As $\|\epsilon\|_2 \to 0$, for a new neighbor $\bx_i$ that was not an old neighbor, and an old neighbor $\bx_j$, we have $d(\bx_i,\bx+\epsilon) \geq r-\|\epsilon\|_2$ and $d(\bx_j,\bx+\epsilon) \leq r+\|\epsilon\|_2$. 
Therefore as $\|\epsilon\|_2 \to 0$, we know:
$$
\lim_{\|\epsilon\|_2 \to 0}\frac{d(\bx_i,\bx_0)}{d(\bx_j,\bx_0)} \geq 1
$$
Therefore we have:
$$
\lim_{\|\epsilon\|_2 \to 0} w_i^{new} \leq w_j^{'}
$$
for any $j \leq n - k$, and 
$$
\lim_{\|\epsilon\|_2 \to 0} w_i^{'} \leq w_j^{'}
$$
for any $i > n-k$ and $j \leq n - k$. Then combining with the result from assumption 1 and Lemma 1, we know for sufficiently small $\epsilon$ and large $n$, we can find an arbitrarily small constant $C_3$ such that, with probability going to 1, we have $\frac{k+k'}{n} \leq C_3\epsilon$. Denote the required sample size by $n_1$ here. Following the same procedure we have shown, we can choose $N>\frac{2n_1}{p_1}$ to guarantee that $n>n_1$, larger than the required sample size. Denote $b=\frac{-\sum_{i=n-k+1}^nw_i^{'}+\sum_{i=1}^{k'}w_i^{new}}{\sum_{i=1}^{n}w_i^{'}}$. Then we can derive:
\begin{align*}
    |\frac{-\sum_{i=n-k+1}^nw_i^{'}+\sum_{i=1}^{k'}w_i^{new}}{\sum_{i=1}^{n}w_i^{'}}| &\leq     |\frac{-\sum_{i=n-k+1}^nw_i^{'}}{\sum_{i=1}^{n}w_i^{'}}| +     |\frac{\sum_{i=1}^{k'}w_i^{new}}{\sum_{i=1}^{n}w_i^{'}}| \\
    & \leq \frac{k}{n} + \frac{k^{'}}{n} = \frac{k+k^{'}}{n} \\
    & \leq C_3\epsilon
\end{align*}

Thus we know $b\leq C_3\epsilon$. We can further derive:
\begin{align*}
|\frac{-\sum_{i=n-k+1}^nw_i^{'}+\sum_{i=1}^{k'}w_i^{new}}{\sum_{i=1}^{n-k}w_i^{'}+\sum_{i=1}^{k'}w_i^{new}}| &=  |\frac{-\sum_{i=n-k+1}^nw_i^{'}+\sum_{i=1}^{k'}w_i^{new}}{\sum_{i=1}^{n}w_i^{'}-\sum_{i=n-k+1}^nw_i^{'}+\sum_{i=1}^{k'}w_i^{new}}| \\
&= |\frac{b}{1+b}| \leq \frac{C_3\epsilon}{1-C_3\epsilon}    
\end{align*}

We further derive the bound for (C):
{\footnotesize
\begin{align*}
(C)&=\frac{\sum_{i=1}^nw_i^{'}\by_i}{\sum_{i=1}^nw_i^{'}}(1-\frac{\sum_{i=1}^nw_i^{'}}{\sum_{i=1}^{n-k}w_i^{'}+\sum_{i=1}^{k'}w_i^{new}})=\frac{\sum_{i=1}^nw_i^{'}\by_i}{\sum_{i=1}^nw_i^{'}}\frac{-\sum_{i=n-k+1}^nw_i^{'}+\sum_{i=1}^{k'}w_i^{new}}{\sum_{i=1}^{n-k}w_i^{'}+\sum_{i=1}^{k'}w_i^{new}} \\
&=f(x)\frac{-\sum_{i=n-k+1}^nw_i^{'}+\sum_{i=1}^{k'}w_i^{new}}{\sum_{i=1}^{n-k}w_i^{'}+\sum_{i=1}^{k'}w_i^{new}} 
\end{align*}
}

Therefore as $\|\epsilon\|_2 \to 0$, we have $\underset{\|\epsilon\|_2 \to 0}{\lim}\|\frac{(C)}{\epsilon}\|_2\leq C_3C_4$. Similarly, we can obtain a bound with $(D)$:
\begin{align*}
    \underset{\|\epsilon\|_2 \to 0}{\lim}\|\frac{(D)}{\epsilon}\|_2&=\|\frac{-\sum_{i=n-k+1}^nw_i^{'}\by_i^{'}+\sum_{i=1}^{k'}w_i^{new}\by_i^{new}}{(\sum_{i=1}^{n-k}w_i^{'}+\sum_{i=1}^{k'}w_i^{new})\epsilon}\|_2 \\
    &\leq C_4\frac{-\sum_{i=n-k+1}^nw_i^{'}+\sum_{i=1}^{k'}w_i^{new}}{(\sum_{i=1}^{n-k}w_i^{'}+\sum_{i=1}^{k'}w_i^{new})\epsilon} \leq C_3C_4
\end{align*}
We also have the gradient bound for $f(\bx)$. We conclude that:
\begin{align*}
\underset{\|\epsilon\|_2 \to 0}{\lim}\|\frac{f(\bx+\epsilon)-f(x)}{\epsilon}\|_2 &\leq \underset{\|\epsilon\|_2 \to 0}{\lim}(\|\frac{(C)+(D)}{\epsilon}\|_2+ \frac{2C_4C_2}{C_1}+\delta) \\
&\leq C_3C_4+C_3C_4+\frac{2C_4C_2}{C_1}+\delta\\
&=2C_4(\frac{C_2}{C_1}+C_3)+\delta 
\end{align*}
In Lemma 3, we have $C_3 \to 0$ as $N \to \infty$, so we can finally remove $C_3$ from the bound:
$$
\underset{c\to 0}{\lim} \|\frac{f(\bx_0+c\hat{\bu})-f(\bx_0)}{c}\|_2 \leq  2C_4(\frac{C_2}{C_1})+\delta
$$
\end{proof}

After deriving the Lipschitz upper bound of neighborhood preserving layer, we compare it with the Lipschitz bound of a fully-connected layer. We know when only one layer is considered, given $\bX \in \mathbb{R}^{n*p}$ and $\by \in \mathbb{R}^{n*d}$, the best fully-connected layer is equivalent to a multi-response regression problem. Denoting $\bW=(\bw^{(1)},\dots,\bw^{(d)})$, we have:
$$
\bw^{(i)}=(\bX^T\bX)^{-1}\bX^T\by_i
$$
This choice of weights can minimize the $\ell_2$ loss in this specific layer, and is the best unbiased linear weight. When a single layer is considered, this is the target weight we should use. The corresponding feed forward function is defined as $f(\bx) = \bW\bx$. To proceed with the analysis, we introduce a set of regularity conditions for $\bx$ and $\by$.
\begin{assumption}
We assume $\bx_i$'s are independently distributed from $\mathbb{P}$ such that $E(\bx)=\textbf{0}$ and $cov(\bx) \lesssim C_5\textbf{I}_{p}$. 
\end{assumption}
$A \lesssim B$ means that $B-A$ is a positive definite matrix. The assumption requires that the distribution of the low dimensional embedding $\by$ is well behaved, and the covariance matrix has eigenvalue upper bound. It holds naturally as long as $\bx$ is bounded.  Further we assume each $\bx^{(i)}$ and $\by_j$ have correlation $r_{ij}$. All these assumptions can also be easily achieved by our neighborhood preserving layer.

\begin{theorem}
When Assumption 3 holds, given $\delta>0$, if $n$ is sufficiently large then $\bW$ satisfies:
$$
\|\bw^{(j)}\|_2 \geq \frac{1}{C_5+\delta}\sqrt{\sum_{i=1}^pD_i^2r_i^2}-\delta
$$
Furthermore, there exists a direction of $\epsilon$ such that:
$$
\underset{c \to 0}{\lim}\|\frac{f(\bx_0+c\epsilon)-f(\bx_0)}{c}\|_2 \geq  \frac{1}{C_5+\delta}\sqrt{\sum_{i=1}^pD_i^2r_i^2}-\delta
$$
where $D_i=sd(\bx^{(i)})sd(y_i)$, is the product of two standard deviations, providing a lower bound on the Lipschitz constant of this fully connected layer.
\end{theorem}
\begin{remark}
Since the fully connected layer is designed to extract features of $\bx$ and pass them to $\by$, $r_{ii}$ should be large. 
\end{remark} 
\begin{proof}
As we have shown, $\bw^{(i)}=(\bX^T\bX)^{-1}\bX^T\by_i$. Thus we have
$$
\|\bw^{(i)}\|_2=\|(\bX^T\bX)^{-1}\bX^T\by_i\|_2
$$
By the covariance bounded eigenvalue assumption, we know $\underset{n\to\infty}{\lim}\frac{1}{n}\bX^T\bX \lesssim C_5\textbf{I}_p$. Thus we can find sufficiently large $n$, such that $\frac{1}{n}\bX^T\bX \lesssim (C_5+\delta)\textbf{I}$. Writing $\bX^T\by_i=(\bX^{(1)^T}\by_i,\dots,\bX^{(p)^T}\by_i) \in \mathbb{R}^{p}$, we have:
\begin{align*}
\|(\bX^T\bX)^{-1}\bX^T\by_i\|_2 &\geq \frac{1}{(C_5+\delta)n}\|\bX^T\by_i\|_2 \\
&\geq \frac{1}{C_5+\delta}\sqrt{\sum_{i=1}^p (cov_n(\bX^{(i)},\by_i))^2}
\end{align*}
We know:
$$
\underset{n\to\infty}{\lim} cov_n(\bX^{(i)},\by_i) \to cov(\bx^{(i)},y_i)=cor(\bx^{(i)},y_i)sd(\bx^{(i)})sd(y_i) = D_ir_i
$$
where $D_i=sd(\bx^{(i)})sd(\by_i)$. Plugging into the previous equation, for any $\delta>0$, we can find a sufficiently large $n$ such that:
\begin{align*}
    \|\bW\|_2 \geq \|\bw^{(i)}\|_2 \geq  \frac{1}{C_5+\delta}\sqrt{\sum_{i=1}^pD_i^2r_i^2} - \delta
\end{align*}

Finally, we have the Lipschitz constant satisfies:
{\footnotesize
$$
\underset{\|\epsilon\|_2\to 0}{\lim}\|\frac{f(\bx_0+\epsilon)-f(\bx_0)}{\epsilon}\|_2 \geq \underset{\bx \in \mathbb{R}^p}{\sup}\frac{\|\bw^{(i)}\bx\|_2}{\|\bx\|_2}=\|\bw^{(i)}\|_2\geq  \frac{1}{C_5+\delta}\sqrt{\sum_{i=1}^pD_i^2r_i^2} - \delta
$$
}
\end{proof}

So far we have derived the Lipschitz upper bound of our neighborhood preserving layer:  $T_1=2C_4(\frac{C_2}{C_1}+C_3)+\delta$, and the lower bound of the fully connected regression layer: $T_2=\frac{1}{C_5+\delta}\sqrt{\sum_{i=1}^pC_i^2r_i^2}-\delta$. We see $\frac{T_1}{T_2}=o(1/\sqrt{p})$ when all $r_i$ are O(1). It means in general our neighborhood layer is on the $o(1/\sqrt{p})$ order of the Lipschitz bound of the fully connected layer.

The derived Lipschitz bound is closely related to the robustness of the network, and also the gradient descent based attack method. If the Lipschitz constant is small overall, then perturbations from all directions cannot significantly change the loss function, and the gradient descent based attack will be ineffective.

To illustrate this effect, first we need to introduce `minimal $L_p$ distortion', which is a well acknowledged metric for robustness evaluation (Hein and Andriushchenko 2017).
\begin{definition}
Let $C$ be the set of samples with label $c$, $l$ be a class, and $f_c(x)$ be the predicted probability that a point $x$ belongs to class $c$. Then we say a network has minimal $L_p$ distortion $\delta_p$ at point $x$, if $\delta_p$ is defined as:
$$
\underset{\delta \in \mathbb{R}^d}{\min}\|\delta\|_p \ \ s.t. \ \underset{l\neq c}{\max} f_l(x+\delta)\geq f_c(x+\delta) \ and \ x \in C
$$
\end{definition}
$\delta_p$ is the maximal distortion $L_p$ norm allowed such that all distortions smaller than this magnitude will not change the classification label. This metric is closely related to the performance of a network against a C\&W attack\citep{carlini2017towards}. In a C\&W attack, we exactly look for a $L_2$ distortion in $S$ that maximizes the difference in loss function. 
\begin{coro}
If the conditions in Theorem 1 and 2 hold, then the upper bound of minimal $L_2$ distortion bound introduced in Hein and Andriushchenko (2017) will be improved by $\frac{T_2}{T_1}$ times by replacing a fully-connected layer with our proposed neighborhood preserving layer.
\end{coro}
\begin{proof}
Assume the Lipschitz constant previous to the dimension reduction layer is $L_a$, and after the dimension reduction layer is $L_b$. Then as analyzed in Szegedy et al.(2013), the Lipschitz constant of the whole network with the proposed neighborhood preserving layer is $L=L_aL_bT_1$, and for a network with a fully-connected layer is $L^{
}=L_aL_bT_2$. Then we plug the Lipschitz bound into Theorem 2.1 of Hein and Andriushchenko (2017), choosing $p=q=2$ and radius to be sufficiently large, to obtain:
$$
\|\delta\| \leq \{\underset{j\neq c}{\min} \frac{f_c(x)-f_j(x)}{L}\}
$$
Thus we obtain an upper bound of the minimal $L_2$ distortion bound:
$$
\delta_{neighbor}=\frac{T_1}{T_2}\delta_{fc}
$$
\end{proof}

So far, we have analyzed how our neighborhood preserving layer helps shrink the Lipschitz constant and thus helps improve the minimal distortion bound. Madry et al. (2017) propose the saddle point problem, which is well recognized as a good measure of the robustness of a network:
$$
\rho=\mathbb{E}_{x,y} [\underset{\delta \in S}{\max} L(\theta,x+\delta,y)]
$$
where $S$ is the feasible region of a small distortion with radius $\epsilon$, and $y$ is the class associated to input point $x$.

An alternate measure of distortion is:
$$
g_{\delta}= \mathbb{E}_{x,y} [\underset{\delta \in S}{\max} L(\theta,x+\delta,y)] - \mathbb{E}_{x,y}[L(\theta,x,y)]
$$
Here we show that taking advantage of the result from Theorem 1, our robustness will also be significantly improved under this metric. 
\begin{theorem}
Suppose all conditions in Theorem 1 are satisfied, and the loss function for classification is chosen as negative log likelihood loss.  Then the network with bottleneck $f(\cdot)$ (a neighborhood preserving layer or a fully-connected layer) has distortion expectation upper bound:
$$
g_{\delta} \leq L_aL_b\epsilon \int_{\mathbb{R}^p} \underset{\delta \in S}{\max} Lip(f(z+L_a\delta))dz
$$
where $Lip(\cdot)$ is the Lipschitz constant at a specific point, and $L_a$ is the Lipschitz constant prior to the bottleneck layer, and $L_b$ is the Lipschitz constant after the bottleneck layer. Under Assumptions 1-3, the distortion bound of fully-connected layer is $\frac{T_2}{T_1}$ times of our neighborhood preserving layer.
\end{theorem}
\begin{proof}
Let $z=t_1(x)$ represent the random variable after taking all layers ahead of the dimension reduction layer, and let $y=t_2(\cdot,i)$ represent all layers behind the dimension reduction layer, returning the output probability for class $i$. By definition, we have:
\begin{align*}
    g_{\delta}&=\mathbb{E}_{x,y} [\underset{\delta \in S_1}{\max} \  t_2(f(t_1(x+\delta)),y))] - \mathbb{E}_{x,y} [t_2(f(t_1(x)),y))]\\
    &\leq L_b(\mathbb{E}_{z} [\underset{\delta \in S}{\max} \  f(z+L_a\delta)] -  \mathbb{E}_{x} [f(t_1(x))) \\
        &= L_b(\mathbb{E}_{z} [\underset{\delta \in S}{\max} \  f(z+L_a\delta)] -  \mathbb{E}_{z} [f(z)])\\
    &\leq L_b\int ([\underset{\delta \in S}{\max} \ f(z+L_a\delta)]-f(z))dz \\
    &\leq L_aL_b\epsilon \int \underset{\delta \in S}{\max} \  Lip(f(z+L_a\delta))dz
\end{align*}
where $S_1$ and $S$ are the feasible regions of a small distortion with radius $\epsilon$ in the input and high dimensional spaces. 
\end{proof}

From Theorem 3, we see that the distortion bound is proportion to its Lipschitz constant bound. The ratio of bounds between the neighborhood preserving and fully connected bottleneck layers is $\frac{T_1}{T_2}$. Considering $\frac{T_1}{T_2}=o(\frac{1}{\sqrt{p}})$, the distortion expectation of our algorithm has much smaller bounds, making our algorithm more robust against adversarial attack. 

\section{Experiments}
\label{sec:Experiments}
In this section, we demonstrate the practical superiority of our proposed layer on benchmark datasets. Explicitly, we compare the performance of our proposed algorithm (neighbor preserving layer) with benchmark CNN models against a PGD attack. The experiments are conducted on two benchmark datasets:
\begin{itemize}
    \item \textbf{MNIST}: handwritten digit dataset, which consists of 60,000 training images and 10,000 testing images. Theses are $28 \times 28$ black and white images in ten different classes.

    \item \textbf{CIFAR10}: natural image dataset, which contains 50,000 training images and 10,000 testing images in ten different classes. These are low resolution three-channel $32 \times 32$ color images.
\end{itemize}

For the MNIST dataset, the encoder is setup as a two-layer CNN with kernel size $5\times 5$ and 5/20 output channels in the first and second layer, and a $2 \times 2$ pooling following each convolutional layer. The encoder leads to a 800-dimensional high-dimensional embedding. For CIFAR10 dataset, we employ VGG16 architecture as the encoder, and it leads to a 512-dimensional high-dimensional embedding. For the classifier, we employ two fully-connected layers with ReLU activation, and then a softmax function to obtain the final 10-dimensional prediction probability for 10 classes. For dimension reduction layers, two type of dimension reduction layers are employed: (1) Fully-connected layers with ReLU activation. (2) Our proposed neighborhood preserving layer.

We consider the strongest white-box PGD attack over two datasets. Explicitly in our experiment, the $\Pi_{\epsilon}$ is considered as $\ell_{\infty}$ projection over the data. We normalize the data to the ranges from 0 to 1. Therefore $\epsilon=0.01$ represents changes up to about 3 pixels, and $\epsilon=0.05$ represents changes up to about 15 pixels, and so forth. In the table, `FC' represents fully-connected dimension reduction layers, `NP' represents proposed neighborhood preserving dimension reduction layers, and `Ref' represents proposed neighborhood preserving layers with only 1000 reference point instead of full datasets. The subscript indicates the output dimension of the layer. We provide a table with $\ell_\infty$ projection attack under different bottleneck layers:
\begin{table}[h]
    \centering
    \begin{tabular}{c|c|c|c|c|c|c}
       Perturbation & $FC_8$ & $FC_{16}$ & $NP_8$ & $NP_{16}$ &$Ref_{8}$ &$Ref_{16}$\\ \hline
        $\epsilon=0.01$ & 98.56  &$\boldsymbol{98.98}$&96.85& 96.95& 94.42&94.52\\
        $\epsilon=0.05$ & 84.78 &88.46&$\boldsymbol{94.82}$&94.26&90.40 &90.95\\
        $\epsilon=0.1$ &37.82 & 46.13&$\boldsymbol{91.67}$&90.72& 81.76&82.69\\
        $\epsilon=0.2$ & 9.67 & 11.30&$\boldsymbol{89.73}$&89.04& 73.89&74.92\\
    \end{tabular}
    \caption{MNIST data set: Accuracy result under $\ell_{\infty}$ projection PGD attack}
    \label{tab:mnist}
\end{table}

\begin{table}[h]
    \centering
    \begin{tabular}{c|c|c|c|c}
       Perturbation & $FC_8$ & $FC_{16}$ & $NP_8$ & $NP_{16}$\\ \hline
        $\epsilon=0.01$ & 66.66  &68.43&76.17&$\boldsymbol{77.76}$\\
        $\epsilon=0.05$ & 19.29 &20.50&65.40&$\boldsymbol{66.10}$\\
        $\epsilon=0.1$ &7.38 & 8.06&61.33&$\boldsymbol{62.04}$\\
        $\epsilon=0.2$ & 5.46 & 5.57&58.84&$\boldsymbol{59.91}$\\
    \end{tabular}
    \caption{CIFAR10 data set: Accuracy result under $\ell_{\infty}$ projection PGD attack}
    \label{tab:cifar10}
\end{table}

From these two tables, we draw several conclusions: (1) Coinciding with our theoretical results, our proposed layer is much more robust against a PGD attack compared with fully-connected layers, especially when $\epsilon$ is large. (2) Comparing the two tables, we can see the true intrinsic dimension of a dataset is important to the performance of networks. The MNIST dataset is known to be a simpler dataset and has smaller intrinsic dimension than CIFAR10. It helps explain why $NP_8$ outperforms $NP_{16}$ in MNIST, while $NP_{16}$ outperforms $NP_8$ in CIFAR10. Our results coincide with both our findings about intrinsic dimension of the two datasets. (3) In the MNIST dataset, we also consider different choices of reference points. We consider using all training data as reference points, versus using 1000 points as reference points. We see that using 1000 reference points sacrifices a bit of prediction performance, while it can significantly reduce the computational burden, reducing $98.3\%$ of reference points. (4) For MNIST data,  $\epsilon=0.01$ is a weak attack, and we see that our model maintain high accuracy under these settings. It demonstrates the general interest of our layer.

Overall in this section, we illustrate that our algorithm can replace dimension reduction layers and effectively improve the robustness of models. It is also compatible with other state-of-art adversarial training methods. 

\section{Conclusion}
In this paper, we propose a novel dimension reduction layer through neighborhood preservation. The proposed layer can replace fully connected layers in general neural nets to improve robustness against adversarial attack. We provide theoretical analysis and empirical experiments to show that the proposed layer enjoys a smaller gradient upper bound, and is more robust against gradient-based adversarial attack. Neural networks with our proposed layers can be efficiently trained, and are also flexible in adapting other adversarial training procedures. 




\bibliographystyle{plainnat} 
\bibliography{references.bib}

\begin{thebibliography}{23}
\providecommand{\natexlab}[1]{#1}
\providecommand{\url}[1]{\texttt{#1}}
\expandafter\ifx\csname urlstyle\endcsname\relax
  \providecommand{\doi}[1]{doi: #1}\else
  \providecommand{\doi}{doi: \begingroup \urlstyle{rm}\Url}\fi

\bibitem[Afsar et~al.(2008)Afsar, Akram, Arif, and Khurshid]{afsar2008pruned}
Fayyaz~A Afsar, MU~Akram, M~Arif, and J~Khurshid.
\newblock A pruned fuzzy k-nearest neighbor classifier with application to
  electrocardiogram based cardiac arrhytmia recognition.
\newblock In \emph{2008 IEEE International Multitopic Conference}, pages
  143--148. IEEE, 2008.

\bibitem[Arif et~al.(2010)Arif, Akram, et~al.]{arif2010pruned}
Muhammad Arif, Muhammad~Usman Akram, et~al.
\newblock Pruned fuzzy k-nearest neighbor classifier for beat classification.
\newblock \emph{Journal of Biomedical Science and Engineering}, 3\penalty0
  (04):\penalty0 380, 2010.

\bibitem[Bengio et~al.(2004)Bengio, Paiement, Vincent, Delalleau, Roux, and
  Ouimet]{bengio2004out}
Yoshua Bengio, Jean-fran{\c{c}}cois Paiement, Pascal Vincent, Olivier
  Delalleau, Nicolas~L Roux, and Marie Ouimet.
\newblock Out-of-sample extensions for lle, isomap, mds, eigenmaps, and
  spectral clustering.
\newblock In \emph{Advances in neural information processing systems}, pages
  177--184, 2004.

\bibitem[Carlini and Wagner(2017)]{carlini2017towards}
Nicholas Carlini and David Wagner.
\newblock Towards evaluating the robustness of neural networks.
\newblock In \emph{2017 ieee symposium on security and privacy (sp)}, pages
  39--57. IEEE, 2017.

\bibitem[Chakraborty et~al.(2018)Chakraborty, Alam, Dey, Chattopadhyay, and
  Mukhopadhyay]{chakraborty2018adversarial}
Anirban Chakraborty, Manaar Alam, Vishal Dey, Anupam Chattopadhyay, and Debdeep
  Mukhopadhyay.
\newblock Adversarial attacks and defences: A survey.
\newblock \emph{arXiv preprint arXiv:1810.00069}, 2018.

\bibitem[Chen et~al.(2018)Chen, Sharma, Zhang, Yi, and Hsieh]{chen2018ead}
Pin-Yu Chen, Yash Sharma, Huan Zhang, Jinfeng Yi, and Cho-Jui Hsieh.
\newblock Ead: elastic-net attacks to deep neural networks via adversarial
  examples.
\newblock In \emph{Thirty-second AAAI conference on artificial intelligence},
  2018.

\bibitem[Dong et~al.(2018)Dong, Liao, Pang, Su, Zhu, Hu, and
  Li]{dong2018boosting}
Yinpeng Dong, Fangzhou Liao, Tianyu Pang, Hang Su, Jun Zhu, Xiaolin Hu, and
  Jianguo Li.
\newblock Boosting adversarial attacks with momentum.
\newblock In \emph{Proceedings of the IEEE conference on computer vision and
  pattern recognition}, pages 9185--9193, 2018.

\bibitem[Goodfellow et~al.(2014)Goodfellow, Shlens, and
  Szegedy]{goodfellow2014explaining}
Ian~J Goodfellow, Jonathon Shlens, and Christian Szegedy.
\newblock Explaining and harnessing adversarial examples.
\newblock \emph{arXiv preprint arXiv:1412.6572}, 2014.

\bibitem[He et~al.(2005)He, Cai, Yan, and Zhang]{he2005neighborhood}
Xiaofei He, Deng Cai, Shuicheng Yan, and Hong-Jiang Zhang.
\newblock Neighborhood preserving embedding.
\newblock In \emph{Tenth IEEE International Conference on Computer Vision
  (ICCV'05) Volume 1}, volume~2, pages 1208--1213. IEEE, 2005.

\bibitem[Hein and Andriushchenko(2017)]{hein2017formal}
Matthias Hein and Maksym Andriushchenko.
\newblock Formal guarantees on the robustness of a classifier against
  adversarial manipulation.
\newblock In \emph{Advances in Neural Information Processing Systems}, pages
  2266--2276, 2017.

\bibitem[Li(2011)]{li2011concise}
Shengqiao Li.
\newblock Concise formulas for the area and volume of a hyperspherical cap.
\newblock \emph{Asian Journal of Mathematics and Statistics}, 4\penalty0
  (1):\penalty0 66--70, 2011.

\bibitem[Maaten and Hinton(2008)]{maaten2008visualizing}
Laurens van~der Maaten and Geoffrey Hinton.
\newblock Visualizing data using t-sne.
\newblock \emph{Journal of machine learning research}, 9\penalty0
  (Nov):\penalty0 2579--2605, 2008.

\bibitem[Madry et~al.(2017)Madry, Makelov, Schmidt, Tsipras, and
  Vladu]{madry2017towards}
Aleksander Madry, Aleksandar Makelov, Ludwig Schmidt, Dimitris Tsipras, and
  Adrian Vladu.
\newblock Towards deep learning models resistant to adversarial attacks.
\newblock \emph{arXiv preprint arXiv:1706.06083}, 2017.

\bibitem[McInnes et~al.(2018)McInnes, Healy, and Melville]{mcinnes2018umap}
Leland McInnes, John Healy, and James Melville.
\newblock Umap: Uniform manifold approximation and projection for dimension
  reduction.
\newblock \emph{arXiv preprint arXiv:1802.03426}, 2018.

\bibitem[Moosavi-Dezfooli et~al.(2016)Moosavi-Dezfooli, Fawzi, and
  Frossard]{moosavi2016deepfool}
Seyed-Mohsen Moosavi-Dezfooli, Alhussein Fawzi, and Pascal Frossard.
\newblock Deepfool: a simple and accurate method to fool deep neural networks.
\newblock In \emph{Proceedings of the IEEE conference on computer vision and
  pattern recognition}, pages 2574--2582, 2016.

\bibitem[Ross and Doshi-Velez(2017)]{ross2017improving}
Andrew~Slavin Ross and Finale Doshi-Velez.
\newblock Improving the adversarial robustness and interpretability of deep
  neural networks by regularizing their input gradients.
\newblock \emph{arXiv preprint arXiv:1711.09404}, 2017.

\bibitem[Roweis and Saul(2000)]{roweis2000nonlinear}
Sam~T Roweis and Lawrence~K Saul.
\newblock Nonlinear dimensionality reduction by locally linear embedding.
\newblock \emph{science}, 290\penalty0 (5500):\penalty0 2323--2326, 2000.

\bibitem[Simonyan and Zisserman(2014)]{simonyan2014very}
Karen Simonyan and Andrew Zisserman.
\newblock Very deep convolutional networks for large-scale image recognition.
\newblock \emph{arXiv preprint arXiv:1409.1556}, 2014.

\bibitem[Srivastava et~al.(2014)Srivastava, Hinton, Krizhevsky, Sutskever, and
  Salakhutdinov]{srivastava2014dropout}
Nitish Srivastava, Geoffrey Hinton, Alex Krizhevsky, Ilya Sutskever, and Ruslan
  Salakhutdinov.
\newblock Dropout: a simple way to prevent neural networks from overfitting.
\newblock \emph{The journal of machine learning research}, 15\penalty0
  (1):\penalty0 1929--1958, 2014.

\bibitem[Szegedy et~al.(2013)Szegedy, Zaremba, Sutskever, Bruna, Erhan,
  Goodfellow, and Fergus]{szegedy2013intriguing}
Christian Szegedy, Wojciech Zaremba, Ilya Sutskever, Joan Bruna, Dumitru Erhan,
  Ian Goodfellow, and Rob Fergus.
\newblock Intriguing properties of neural networks.
\newblock \emph{arXiv preprint arXiv:1312.6199}, 2013.

\bibitem[Szegedy et~al.(2017)Szegedy, Ioffe, Vanhoucke, and
  Alemi]{szegedy2017inception}
Christian Szegedy, Sergey Ioffe, Vincent Vanhoucke, and Alexander~A Alemi.
\newblock Inception-v4, inception-resnet and the impact of residual connections
  on learning.
\newblock In \emph{Thirty-first AAAI conference on artificial intelligence},
  2017.

\bibitem[Tenenbaum et~al.(2000)Tenenbaum, De~Silva, and
  Langford]{tenenbaum2000global}
Joshua~B Tenenbaum, Vin De~Silva, and John~C Langford.
\newblock A global geometric framework for nonlinear dimensionality reduction.
\newblock \emph{science}, 290\penalty0 (5500):\penalty0 2319--2323, 2000.

\bibitem[Weng et~al.(2018)Weng, Zhang, Chen, Yi, Su, Gao, Hsieh, and
  Daniel]{weng2018evaluating}
Tsui-Wei Weng, Huan Zhang, Pin-Yu Chen, Jinfeng Yi, Dong Su, Yupeng Gao,
  Cho-Jui Hsieh, and Luca Daniel.
\newblock Evaluating the robustness of neural networks: An extreme value theory
  approach.
\newblock \emph{arXiv preprint arXiv:1801.10578}, 2018.

\end{thebibliography}







\end{document}